\documentclass[review]{elsarticle}

\usepackage{paralist}
\usepackage{graphicx}
\usepackage{comment}
\usepackage{color}
\usepackage{subfig}
\usepackage{hyperref}

\journal{Robotics and Autonomous Systems}

\usepackage{amssymb}

\usepackage{amssymb}
\usepackage{amsmath}
\usepackage{xcolor}
\usepackage{multirow}
\usepackage{booktabs}
\usepackage[ruled,vlined]{algorithm2e}

\SetKwInput{KwInput}{Input}
\SetKwInOut{KwOutput}{Output}
\renewcommand{\vec}[1]{\boldsymbol{#1}}

\newcommand{\mat}[1]{\boldsymbol{#1}}

\DeclareMathOperator*{\argmax}{arg\,max}

\usepackage{soul}

\newcommand\alWuline{\bgroup\markoverwith{\textcolor{orange}{\rule[-0.5ex]{2pt}{1pt}}}\ULon}

\begin{document}

\begin{frontmatter}

\title{Benchmarking 6D Object Pose Estimation for Robotics}
\title{6D Object Pose Estimation for Robotics Revisited: Unified Benchmark Dataset and Performance Metric}
\title{Object Pose Estimation in Robotics Revisited}

\author[1,2]{Antti Hietanen\corref{cor1}%
}%

\author[2]{Jyrki Latokartano
}%

\author[1]{Alessandro Foi}
\author[2]{Roel Pieters}
\author[3]{Ville Kyrki}
\author[2]{Minna Lanz}
\author[1]{Joni-Kristian K\"am\"ar\"ainen}

\address[1]{Computing Sciences, Tampere University, Finland}
    \address[2]{Automation Technology and Mechanical Engineering, Tampere University, Finland}
\address[3]{Department of Electrical Engineering and Automation, Aalto University, Finland}

\cortext[cor1]{Corresponding author. This project has received funding from the European Union's Horizon 2020 research and innovation programme under grant agreement No 825196.}

\begin{abstract}
Vision based object grasping and manipulation in robotics require accurate estimation of object's 6D pose. 
The 6D pose estimation has received significant attention in
computer vision community and multiple datasets and evaluation metrics have been proposed.
However, 
the existing metrics measure how well two geometrical surfaces are aligned - ground truth vs. estimated pose - which does not directly measure how well a robot can perform the task with the given estimate.
In this work we propose a probabilistic metric that directly measures success in robotic tasks.
The evaluation metric is based on non-parametric probability density that is estimated from samples of a real physical
setup. During the pose evaluation stage the physical setup is not needed.
The evaluation metric is validated in controlled experiments and a new pose estimation dataset of industrial parts is introduced. The experimental results with the parts confirm that the proposed evaluation metric better reflects the true performance in robotics than the existing metrics.

\end{abstract}


\end{frontmatter}

\section{Introduction}\label{sec:introduction}
One of the most common application in robotics is object manipulation where the fundamental task is to interact with objects in the environment.
Succeeding in a such task requires accurate positioning of the robot end effector respect to the object, especially when interacting with objects having complex shape.
In the literature, lot of works have focused on identifying and generating robust grasp pose hypothesis around a previously unseen object. 
Most of the recent methods rely on learning-based techniques, such as Convolutional Neural Networks (CNN) \cite{do2018affordancenet, gualtieri2016high, pinto2016supersizing, redmon2015real}, which allow learning of features from visual input that correspond to good quality grasps.
More similarly to our work, probabilistic frameworks for grasp pose detection has been proposed in \cite{detry2011learning} where the grasp affordance model is generated by trial-and-error exploration and using the geometric properties of the 3D object.
However, getting the object from the bin into the gripper in some manner does not guarantee successful precision manipulation or wrenching.
Moreover, in industrial assembly, the objects are known before hand and the whole task is implemented based on a single object-related grasp pose which is selected by an experienced engineer. 
In this scenario, the estimated 6D pose of an object has the biggest contribution to grasp quality and eventually to whole task attempt.

Vision-based object recognition and 6D pose estimation from
RGB-D input have
recently become an active research topic in computer vision~\cite{Manhardt_2018_ECCV,SSD6D}. In a typical workflow, a method first recognizes the object in a scene using RGB input and then estimates and refines the
pose using depth (D) which provides a 3D point cloud; the method output is a 6D object pose with respect to the world coordinate frame.
The methods are trained and optimized using training samples with ground truth pose annotations. Several 6D pose estimation datasets have been recently proposed~\cite{hodan2018bop,yang2017performance,hodavn2016evaluation,hinterstoisser2012model} for method comparison.
The two most popular performance metrics are
{\em average absolute translation/orientation error} and
\textit{{\em average distance of corresponding model points (ADC)}}, calculated using the
ground truth and estimated poses. A significant limitation of these metrics is that they effectively measure only the difference between two transformation matrices but this difference is not necessarily indicative of the success in any
particular task with a real robot. 
The robot vision community benefits from datasets and evaluation metrics that measure the actual success in real tasks without requiring physical setups.

The present work aims at providing a proper evaluation metric for robotic pose estimation and a demonstration dataset constructed using the proposed metric and required procedures. Specifically, we introduce a new benchmark dataset and performance metric for evaluating 6D pose estimation methods in robotics. 
The proposed benchmark does not require replication of the physical setup and yet it provides performance numbers valid for real tasks on real setups. 
The benchmark dataset consists of 3D models of industry relevant objects and approximately $600$ test scenes with various amounts of clutter and occlusions.
The provided performance metric is based on a conditional probability model that encodes the properties of the particular assembly task and measures the success in the task for estimated object poses.   

\vspace{\medskipamount}\noindent\textbf{The main contributions} of this work are:
\begin{compactitem}
\item A statistical formulation of a successfully conducted robotic task ($X\!=\!1$) given the estimated object pose.
Concretely, the estimated object pose is converted and parametrized as 6D pose $\vec{\hat{\theta}}$ of the robot gripper  in the object-relative coordinate space and evaluated using a conditional probability metric~$P(X\!=\!1 | \vec{\hat{\theta}})$.
Interpretation of the metric is intuitive: $0.9$ means that on average ninety out of one hundred attempts succeed with the given pose estimate.
The 6D pose vector~$\vec{\hat{\theta}}$
belongs to the 6D space
${\mathcal{E}\!=\!\mathbb{R}^3\!\times\!{S}^3}$, where $S^3$ denotes the 3D sphere parametrized by hyperspherical coordinates. 
Practical grasp probabilities are computed
using non-parametric kernel regression on a number of collected random samples in $\mathcal{E}$ with the physical setup.

\item
An algorithm to generate automatically a large number of random samples for estimating the evaluation probabilities. The algorithm is validated with several real setups where random samples are generated
by a robot arm in industrial assembly tasks and with different
  grippers and objects. Sample success
  or failure ($X\!=\!\left\{0,1\right\}$) is automatically detected to generate thousands of samples in 24 hours (video example \footnote{\url{https://youtu.be/g4e_-p4fTEI}}).
  
\item A public benchmark for 6D object pose estimation in robotics. The benchmark consists of object models and test scenes with
ground truth pose annotations and pre-computed probability models for each task configuration.
In the experimental part of the work, the benchmark is used to
evaluate several baseline and recent pose estimation methods.
\end{compactitem}
It should be noted that the users of our
benchmark do not need the physical setups to evaluate their methods and all performance numbers are still valid for the real setup in our laboratory. On the other hand, the proposed sampling procedures can be used to construct novel benchmarks with different physical setups in other laboratories.
All code and data will be made publicly available to facilitate fair
comparisons and to promote pose estimation research in robotics.

\section{Related Work}
\label{sec:related}
Section~\ref{sec:benchmarks} provides a brief review of the existing pose estimation datasets and their performance metrics, and Section~\ref{sec:SotA} introduces popular baseline and more recent algorithms for 6D object pose estimation.

\subsection{Benchmark datasets and performance metrics}
\label{sec:benchmarks}
Our main focus is on pose estimation from 3D data which is today easily available due to good quality and inexpensive RGB-D (color + depth) sensors.
The authors acknowledge
that there are numerous works dealing with 3D recognition and pose estimation from 2D input such as gray level or color images. There are also many available "2D-to-3D" benchmark
datasets such as Pascal3D~\cite{Xiang-2014-wacv}. However, for practical robot manipulation RGB is often too limited setting and 3D sensing can
be readily adopted.

The early works did evaluations on 3D object models from 3D scan datasets and synthetic scenes. A popular dataset is the Stanford 3D Scanning Repository~\cite{Turk-1994-siggraph} that contains the famous Stanford Bunny. For these datasets the typical performance
metrics are {\em 3D translation and 3D rotation errors}~\cite{shotton2013scene}.

One of the first real and still widely used 3D object
recognition and 6D pose estimation datasets is LineMod introduced by Hinterstoisser et al.~\cite{hinterstoisser2012model}. LineMod training data consists of
3D models and turn-table captured RGB and depth images. The test data consists of various cluttered scenes that were capture from multiple view points on a turn-table. As a unified performance metric
Hinterstoisser et al. proposed to use the ADC metric which calculates the distance between model points transformed by the ground truth pose and the estimated pose. The metric is intuitive as it directly evaluates the fit of the two surfaces. 
However the metric is not well defined for objects having symmetric properties. Hinterstoisser proposed a metric
where the distance between the corresponding points were replaced with the distance to the closest point and thus avoiding the symmetry problem.

Recently, Hodan et~al.~\cite{hodan2018bop} introduced Benchmark for 6D Object Pose Estimation (BOP). BOP contains eight similarly captured datasets, including LineMod, that span various kinds of objects and scenes from household objects and scenes~\cite{doumanoglou2016recovering} to industrial~\cite{hodan2017tless}. Hodan et al.
evaluated 15 recent methods on all eight datasets using a unified
evaluation protocol. Their evaluation protocol takes into account
view point dependent pose uncertainty and therefore they adopted
the {\em Visible Surface Discrepancy} (VSD)~\cite{hodavn2016evaluation}
as the main error metric.
VSD is invariant to pose ambiguity, i.e.~due to the object symmetry there can be multiple poses that are indistinguishable. 
However the method requires additional ground truth in the form of visibility masks.

All above datasets and metrics measure the pose error as the misalignment between
the ground truth and estimated object surface points. This requirement is important, for example, in augmented reality applications where the perceived virtual object must align well with the real environment. However, in robotics the performance metric should
measure success in the target tasks such as
industrial assembly or disassembly.

\subsection{6D pose estimation methods}
\label{sec:SotA}
Sate-of-the-art methods divide RGB-D object pose estimation into
two stages~\cite{SSD6D,Manhardt_2018_ECCV}:
i) detection of objects from RGB and ii) detected object pose estimation
from depth (point cloud). Object detection is out of the scope of this work
and therefore we briefly discuss the methods in
the recent evaluation by Yang~et~al.~\cite{yang2017performance}
with their codes available
(note that NNSR is a robustified version of SS).

\paragraph{Random Sample Consensus (RANSAC)}
RANSAC is a widely used technique for 6D pose estimation~\cite{fischler1981random,brachmann2014learning,Brachmann-2016-cvpr} adopted from the 2D domain. %
It is an iterative process
that uses random sampling technique to generate candidate
solutions for a model (transformation) that aligns two surfaces
with a minimum point-wise error. Free parameter of the
method is $N_{\text{RANSAC}}$ which is the maximum count of pose hypothesis the algorithm samples matches from the correspondence set.
The algorithm iteratively samples candidate transformations which are evaluated by transforming all the matched points and calculating the Euclidean distance
between the corresponding points. All transformed point matches with
distance less than $d_{\text{RANSAC}}$ are counted as inliers. The final
pose is estimated using all inlier points for transformation with the
largest number of inliers.

\paragraph{Hough Transform (HG)} Hough transform~\cite{hough1962method} is an alternative
to RANSAC; instead of random samples each point match
casts votes and pose with
the largest number of votes is selected. There are several methods adopting
this principle~\cite{Knopp-2010-eccv,tombari2010object} and for
the experiments the Hough Grouping (HG) method by Tombari et al.~\cite{tombari2010object} was selected. For fast computation,
the  method  uses a  unique  model reference point (mass centroid) and bins represent pose around the reference point. To  make  correspondence  points  invariant  to  rotation and translation  between  the  model  and  scene,  every  point  is associated  with  a local reference  frame~\cite{tombari2010unique}. The main
parameter of the method is the pose bin size - coarse size provides faster computation but increases pose uncertainty.

\paragraph{Spectral Technique (ST)} 
Leordeanu and Hebert~\cite{leordeanu2005spectral} proposed a spectral grouping technique to find coherent clusters from the initial set of feature matches.
The method takes into account the relationship between points and correspondences and finally uses an eigen-decomposition to estimate the confidence of a correspondence to be an inlier.

First the algorithm creates an affinity matrix $\mat{M}$ which entries represent weighted links between correspondences.
The weights are estimated by calculating the pairwise similarity between two correspondences using a rigidity constraint:
\begin{equation}\label{eq:rigid}
M({c_i},{c_j}) = \min \left\{ {\frac{{{{\| {{\vec{x}_i} - {\vec{x}_j}} \|}}}}{{{{\| {\vec{x}'_i - \vec{x}'_j} \|}}}},\frac{{{{\| {\vec{x}'_i - \vec{x}'_j} \|}}}}{{{{\| {{\vec{x}_i} - {\vec{x}_j}} \|}}}}} \right\}\,,
\end{equation}
where $\vec{x}$ and $\vec{x}'$ are the model and captured scene 3D points, respectively. 
The diagonal elements of the matrix measure the level of individual assignments i.e.~how well $f_i$ and $f_i^{'}$ match.
After computing $\mat{M}$, the principle eigenvector $\vec{v}$ of $\mat{M}$ is calculated and the location of the maximum value $v_i$ gives the highest confidence of $c_i$ being in the inlier set. Next, all the correspondences conflicting with $c_i$ are removed from the initial set of matches $\mat{C}$ and procedure is repeated until $v_i=0$ or $\mat{C}$ is empty and finally the generated inlier set is returned.

\paragraph{Geometric Consistency (GC)} 
While the RANSAC and Hough transform
based methods operate directly on the 3D points there
are methods that exploit the local neighborhood of points to establish more
reliable matches between model and scene point
clouds~\cite{GC,glent2014search}. Geometric Consistency Grouping (GC)~\cite{GC} is a strong baseline and it has been implemented in several point cloud libraries. 
GC works independently from the feature space and utilizes only the spatial relationship of the corresponding points.
The algorithm evaluates the consistency of two correspondences $c_i$ and $c_j$ using a compatibility score
\begin{equation} \label{eq:GC}
d(c_i, c_j) = \Bigl|\left\Vert \vec{x}_{i} - \vec{x}_{j} \right\Vert - \left\Vert \vec{x}_{i}' - \vec{x}_{j}' \right\Vert \Bigl| < \tau_{\text{GC}} \enspace.
\end{equation}
GC simply measures distances near the points and
assigns correspondences to the same cluster if their geometric inconsistency is smaller than the threshold value $\tau_{\text{GC}}$. 

GC is initialized with a fixed number of clusters each having a seed correspondence. Then for each cluster it iteratively searches correspondences which satisfy the compatibility score \eqref{eq:GC}, mark them as visited and continue the process until all the correspondences are visited. Finally, all the cluster sets can be optionally refined using RANSAC.  
In principle, the GC algorithm can return more than one cluster and for pose estimation the cluster with the largest number of correspondences is used
as the pose estimate \cite{Hietanen-icra-2017}. 

\paragraph{Search of Inliers (SI)} 
A recent method by Buch et al.~\cite{glent2014search} achieves state-of-the-art
on several benchmarks. It uses two consecutive processing stages, local voting and global voting. The  first  voting  step  performs local  voting,  where  locally  selected  correspondence  pairs are  selected  between a model and scene,  and  the  score  is computed using their pair-wise similarity score $s_L(c)$.
At the global voting stage, the algorithm samples point correspondences, estimates a transformation and gives a global score to the points correctly aligned outside the estimation point set: $s_G(c)$. The final score $s(c)$~is computed by combining the local and global scores, and finally $s(c)$ are thresholded to inliers and outliers based on Otsu's bimodal distribution thresholding.

\section{Evaluating Object Pose in Robotic}
The standard procedure in industrial
robotics is to manually set up and program the needed manipulation task. An experienced engineer is able
to find a stable pose for grasping and select a gripper and fingers that are good for the given
task. However, all settings are made with the assumption
that object pose is accurate but which is difficult to
achieve in practice even with the best computer vision methods.

A probabilistic formulation of success in the given task with a pose estimate is derived in
Section~\ref{sec:probability}. This formulation is used to define sampling procedures to construct a pose estimation benchmark 
for a physical setup (task) in
Section~\ref{sec:sampling}. However, the users of
a benchmark do not need the physical setup but only
a set of test images, pose ground truth and the estimated
probability function.
\subsection{Probability of completing a programmed task $P(X=1)$} 
\label{sec:probability}
The success of a robot to complete its task is a binary random variable $X\in\left\{0,1\right\}$ 
where $X\!=\!1$ denotes a successful attempt and $X\!=\!0$ denotes an unsuccessful
attempt (failure). Therefore, $X$ follows the Bernoulli distribution,
$P(X|p)=p^X(1-p)^{1-X}$, with complementary probability of success and failure: $E(X)=P(X\!=\!1) =1-P(X\!=\!0)$, where $E$ denotes the mathematical expectation.
The pose is defined by 6D pose coordinates 
$\vec{\theta} = (t_x, t_y, t_z, r_x, r_y, r_z)^T$ where the origin
is the object centric coordinate frame.
The translation vector $(t_x,t_y,t_z)^T\!\in\!\mathbb{R}^3$ and 3D rotation $(r_x,r_y,r_z)^T\!\in\!SO(3)$ both have three degrees of freedom.  The rotation is in axis-angle representation, where the length of the 3D rotation vector is the amount of rotations in radians, and the vector itself gives the axis about which to rotate.
Adding pose to the formulation makes the
success probability a conditional distribution and expectation a conditional
expectation. 
The conditional probability of a successful attempt is
\begin{equation}
 p\left(\vec{\theta}\right) = E(X | \vec{\theta} )=P(X\!=\!1 | \vec{\theta} ) = 1-P(X\!=\!0|\vec{\theta}) \enspace.
\end{equation}

The maximum likelihood
estimate of the Bernoulli parameter $p\in\left[0,1\right]$ from $N$ homogeneous samples $y_i$, $i\!=\!1,\dots,N$, is the sample average
\begin{equation}
  \hat{p}_{\text{ML}} = \frac{1}{N} \sum_{i=1}^N y_i \enspace,
  \label{eq:ml}
\end{equation}
where homogeneity means that all samples are realization of a common Bernoulli random variable with unique underlying parameter $p$. However, guaranteeing homogeneity would require that the samples $\left\{y_i, i\!=\!1,\dots,N\right\}$ were either all collected at the same pose 
$\vec{\theta}_1\!=\dots=\!\vec{\theta}_N$, or for different poses that nonetheless yield same probability $p(\vec{\theta}_1)\!=\dots=\!p(\vec{\theta}_N)$, i.e.~it would require us either to collect multiple samples for each $\vec{\theta}\!\in\!SE(3)$ or to know beforehand $p$ over $SE(3)$ (which is what we are trying to estimate). 
This means that in practice $p$ must be estimated from non-homogeneous samples, i.e.~from  $\left\{y_i, i\!=\!1,\dots,N\right\}$ sampled at pose $\left\{\vec{\theta}_i,  i\!=\!1,\dots,N\right\}$ which can be different and having different underlying  $\left\{p(\vec{\theta}_i),  i\!=\!1,\dots,N\right\}$.

The actual form of $p$ over $SE(3)$  is unknown and
depends on many factors, e.g., the shape of an object, properties of a gripper and a task to be completed. Therefore it is not meaningful to
assume any parametric shape such as the Gaussian or uniform
distribution. Instead, we adopt the Nadaraya-Watson non-parametric estimator which gives the {\em probability of a successful attempt} as
\begin{equation}
\label{eq:prob_grasp}
  \hat{p}_{\vec{h}}\!\left(\vec{\theta}\right) = \frac{\sum_{i=1}^N y_i K_{\vec{h}}\!\left({\vec{\theta}_i-\vec{\theta}}\right)}
 {\sum_{i=1}^N K_{\vec{h}}\!\left({\vec{\theta}_i-\vec{\theta}}\right)} \enspace ,
\end{equation}
where $\vec{\theta}_i$ denotes the poses at which $y_i$ has been sampled and $K_{\vec{h}}:\mathcal{E}\rightarrow\mathbb{R}^+$ is a non-negative multivariate kernel with vector scale ${{\vec{h}}=\left(h_{t_x},h_{t_y},h_{t_z},h_{r_x},h_{r_y},h_{r_z}\right)^T\!>\!0}$.

In this work, $K_{\vec{h}}$ is the multivariate Gaussian kernel
\begin{align}
  K_{\vec{h}} \!\left(\vec{\theta}\right) &= 
  G\!\big({\textstyle{\frac{t_x}{h_{tx}}}}\big)\,
  G\!({\textstyle{\frac{t_y}{h_{ty}}}}\big)\,
  G\!({\textstyle{\frac{t_z}{h_{tz}}}}\big)
  \sum_{j\in\mathbb{Z}}G\!\big({\textstyle{\frac{{r_x}+2j\pi}{h_{rx}}}}\big)\, \cdot\qquad\ \notag\\
  \ &\qquad\quad\,\sum_{j\in\mathbb{Z}}G\!\big({\textstyle{\frac{{r_y}+2j\pi}{h_{ry}}}}\big)
  \sum_{j\in\mathbb{Z}}G\!\big({\textstyle{\frac{{r_z}+2j\pi}{h_{rz}}}}\big)\,,\label{eq:G6D}
\end{align}
where $G$ is the standard Gaussian bell, \smash{$G\left(\theta\right)=\left(2\pi\right)^{-\frac{1}{2}}e^{\frac{1}{2}\theta^2}$}. The three sum terms in \eqref{eq:G6D} realize the modulo-$2\pi$ periodicity of $SO(3)$.

The performance of the estimator \eqref{eq:prob_grasp} is heavily affected by the choice of $\vec{h}$, which determines the influence of samples $y_i$ in computing ${\hat{p}}_{\vec{h}}\!\left(\vec{\theta}\right)$ based on the difference between the estimated and sampled poses 
$\vec{\theta}$ and $\vec{\theta}_i$. Indeed, the parameter $\vec{h}$ can be interpreted as reciprocal to the bandwidth of the estimator: too large $\vec{h}$ results in excessive smoothing whereas too small results in localized spikes.

To find an optimal $\vec{h}$, we use the leave-one-out (LOO) cross-validation method. %
Specifically, we construct the estimator on the basis of $N\!-\!1$ training examples leaving out the $i$-th sample:
\begin{equation}
  \hat{p}^{_\textrm{LOO}}_{\vec{h}}\!\left(\vec{\theta},i\right) = \frac{\sum_{j\neq i} y_j K_{\vec{h}}\!\left({\vec{\theta}_j-\vec{\theta}}\right)}
 {\sum_{j\neq i} K_{\vec{h}}\!\left({\vec{\theta}_j-\vec{\theta}}\right)} \enspace .\notag
 \end{equation}
The likelihood of $y_i$ given $\hat{p}^{_\textrm{LOO}}_{\vec{h}}\!\left(\vec{\theta}_i,i\right)$ is either $\hat{p}^{_\textrm{LOO}}_{\vec{h}}\!\left(\vec{\theta}_i,i\right)$ if $y_i\!=\!1$, or ${1^{\!}-^{\!}\hat{p}^{_\textrm{LOO}}_{\vec{h}}\!\left(\vec{\theta}_i,i\right)}$ if ${y_i\!=\!0}$.
We then select $\vec{h}$ that maximizes the total LOO log-likelihood over the whole set $S_y$:
\begin{equation}
\hat{\vec{h}} = \argmax_{\vec{h}}\! \sum_{i|y_i=1\!}\!\!\log \left( \hat{p}^{_\textrm{LOO}}_{\vec{h}}\!\left(\vec{\theta}_i,i\right)\right)
+\!\!\sum_{i|y_i=0\!}\!\!\log  \left(1\!-\!\hat{p}^{_\textrm{LOO}}_{\vec{h}}\!\left(\vec{\theta}_i,i\right)\right).\notag
\end{equation}

Our choices of the kernel and LOO optimization of the kernel parameters result to probability estimates that are verifiable by controlled experiments (as illustrated in Fig.~\ref{fig:pull_figure}).

\subsection{Sampling the pose space}
\label{sec:sampling}
Section~\ref{sec:probability} provides us a formulation
of the probability of successful robotic manipulation given the object relative grasp pose
$P(X=1|\vec{\theta})$. The practical realization of the probability
values is based on Nadaraya-Watson non-parametric kernel
estimator that requires a number of samples in various poses
$\vec{\theta}_i$ and information of success $y_i=1$ or failure
$y_i=0$ for each attempt.
In this stage, a physical setup is needed for sampling, but the users of the benchmark do not need to replicate the setup - they need only the pre-computed probability densities provided with the benchmark.
For practical reasons we make the following assumptions:
\begin{itemize}
    \item We define a canonical grasp pose respect to a manipulated object %
which is select based on the object intrinsic parameters (i.e.~the distribution of mass) and task requirements (i.e.~on which way the object is being installed).
During the sampling procedure the canonical pose is located using a 2D marker.
    \item We sample the pose space around the canonical grasp pose, %
    and therefore $\vec{\theta}=(t_x,t_y,t_z,r_x,r_y,r_z)^T$ defines $SE(3)$ "displacement" from the canonical grasp pose.
    Sampling was started by first finding the sampling limits of each dimension and then sampling within the limits. %
The limits were found by manually guiding the end effector away from the canonical grasp pose along each dimension until the task execution always failed. The limits are listed in Table~\ref{tab:limits}.

\end{itemize}
With the help of these assumptions we are able to define a sampling procedure that can record samples and their success or failures automatically.
The main limitation of this approach is that the pose space is sampled only near the canonical grasp pose 
which is not guaranteed to be %
the best option in every scenario. 
For instance, the grasp pose might be unreachable due to robot’s kinematic constraints or obstructing objects.
In our work, we assume that the canonical grasp pose is always reachable.

\paragraph{Coordinate transformations}
In the work, a coordinate transformation $\mat{T}_{B}^{A}$ denotes a $4 \times 4$ homogeneous transformation matrix that describes the position of the frame B origin and the orientation of its axes, relative to the reference frame A. 

\begin{figure}[h]
	\begin{center}
		\includegraphics[width=.99\linewidth]{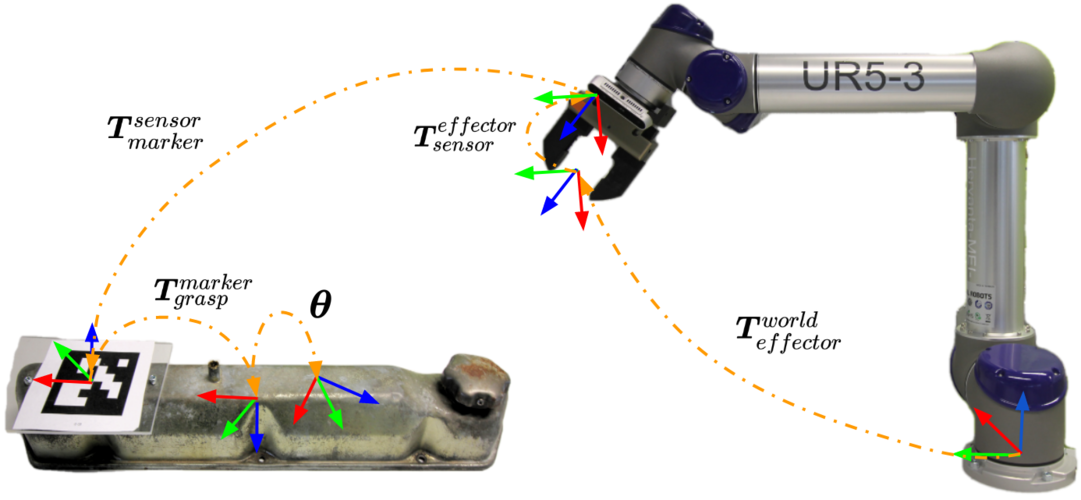}  
		\caption{Coordinate frames used in random sampling of poses for assembly tasks.
			\label{fig:frames}}
	\end{center}
\end{figure}

For a practical implementation used in our experiments the transformation components are (Figure~\ref{fig:frames}):
\begin{compactitem}
\item $\mat{T}_{\textit{grasp}}^{\textit{marker}}$\ \ --\ \ a constant transformation from the canonical grasp pose to the marker frame;
\item $\mat{T}_{\textit{marker}}^{\textit{sensor}}$\ \ --\ \ computed transformation from the marker frame to the sensor frame;
\item $\mat{T}_{\textit{sensor}}^{\textit{effector}}$\ \ --\ \ a constant transformation from the sensor frame to
  the robot end effector frame (camera is attached to the end effector);
\item $\mat{T}_{\textit{effector}}^{\textit{world}}$\ \ --\ \ computed transformation from the end effector frame to the world frame (robot origin).
\end{compactitem}
The world frame is fixed to the robot frame (i.e.~center of the robot base) and programming is based on the tool point that is the end effector frame. 
The coordinate transformation $\mat{T}_{\textit{effector}}^{\textit{world}}$ can be
automatically calculated using the joint angles and known kinematic equations. 
$\mat{T}_{\textit{sensor}}^{\textit{effector}}$ is computed using the standard procedure for
hand-eye calibration with a printed chessboard pattern~\cite{park_robot_1994}. 
Automatic and accurate estimation of the object pose during the sampling is realized by attaching an artificial 2D markers to the manipulated
objects (see Fig.~\ref{fig:setup} for an example). 
For a calibrated camera the ArUco library~\cite{garrido2014automatic} provides an accurate real-time pose of the marker with respect to the
sensor frame $\mat{T}_{\textit{marker}}^{\textit{sensor}}$. 
The constant offset $\mat{T}_{\textit{grasp}}^{\textit{marker}}$ from the marker to the actual grasp pose is object-marker specific and it is estimated manually by hand-guiding the end effector to the desired grasp location on the object (canonical grasp pose) and then measuring the difference between this pose and the marker pose:
\begin{equation}
\mat{T}_{\textit{grasp}}^{\textit{marker}} = \left(\mat{T}_{\textit{marker}}^{\textit{world}}\right)^{-1} \mat{T}_{\textit{grasp}}^{\textit{world}}.\notag
\end{equation}
During the sampling procedure, the canonical grasp pose is then calculated respect to world frame as:
\begin{equation}
\mat{T}_{\textit{grasp}}^{\textit{world}}=
\mat{T}_{\textit{effector}}^{\textit{world}} \cdot
\mat{T}_{\textit{sensor}}^{\textit{effector}} \cdot
\mat{T}_{\textit{maker}}^{\textit{sensor}} \cdot
\mat{T}_{\textit{grasp}}^{\textit{marker}} \enspace
\end{equation}

Finally, samples around the canonical grasp pose are generated from
\begin{equation}
    \hat{\mat{T}}_{\textit{grasp}}^{\textit{world}} = \mat{T}_{\textit{grasp}}^{\textit{world}} \cdot  \mat{\Phi}(\vec{\theta})
    \enspace 
\end{equation}
where the operator $\mat{\Phi}(\cdot)$ converts the 6D pose vector to a $4\times4$ matrix representation
\begin{equation}
    \mat{\Phi}(\vec{\theta}) =
        \begin{bmatrix}
    \mat{R}_{3 \times 3} & \vec{t} \\
    \vec{0} & 1
    \end{bmatrix}.
\end{equation}
The generated pose sample is defined in the vicinity of the canonical pose by the translation shift 
$\vec{t}=(t_x,t_y,t_z)^T$ and rotation matrix $\mat{R} \in \mathbb{R}^{3\times3}$ constructed from the axis-angle vector $(r_x,r_y,r_z)^T$.

\paragraph{Detection of failures}
Each of the manipulated object has a predefined position and orientation how it should be installed, i.e.~\textit{ground truth installation pose}, respect to the target object.
For instance most of the motor parts have to be placed on the motor block precisely in order to fasten the parts with screws.
The robot task is to bring the part to this pose and finally release the part by opening the gripper fingers.
In addition, using excessive force during the task can cause damage to the manipulated objects.
In the work there are two sources of information for detecting manipulation failures:
\begin{itemize}
\item too large difference between the installation pose of the manipulated object and the corresponding ground truth and
\item too large wrench torque at the end effector at any moment of task execution (e.g. due to collisions), including grasping, carrying and installation.
\end{itemize}
Thresholds for the above are task specific and in our experiments they were manually set based on preliminary experiments.

For evaluation the success of the part installation in the terms of correct location the two thresholds are used: $\tau_t$ for the maximum translation error and $\tau_r$ for the maximum orientation error (both task specific). These are computed using installation pose
 $\mat{\hat{\Gamma}}=\big[\hat{\mat{R}}~|~\hat{\vec{t}}\big]$~measured using the marker attached to the manipulated object and the ground truth installation pose $\mat{\Gamma}=\left[\mat{R}~|~\vec{t}\right]$.
Both are measured respect to the target object on which the manipulated object is being installed.
The installation was successful if
\begin{equation}
\begin{split}
&\left\Vert \mat{t} - \hat{\vec{t}} \right\Vert \le \tau_t \\
&\textrm{arccos}\bigg(  \frac{\text{trace} \big(  \hat{\mat{R}}\mat{R}^{-1}     \big) -1}{2}  \bigg)
\le \tau_r
\end{split} \enspace .
\label{eq:verif}
\end{equation}%
The torque is used to detect if the robot collides with its environment during
the task execution. In addition, if the robot places the object to the correct position with too high wrench the whole task is considered as an unsuccessful attempt. The external wrench is computed based on the error between the joint torques required to stay on the programmed trajectory and the expected joint
torques. The robot's internal sensors provide the torque measurements
$\mat{F}\!=\!(f_x, f_y, f_z)$, where $f_x$, $f_y$ and $f_z$ are the forces in
the axes of the robot frame coordinates and measured in Newtons. For each task the limit $f_{max}$ was manually set for each operation stage using preliminary experiments and violating the threshold, i.e
$\left\Vert \mat{F} \right\Vert > f_{max}$, was recorded as failure. All sampling steps are in
Algorithm~\ref{alg:sampling}.
\begin{algorithm}[H]
	\KwInput{Robot program waypoints $\mathcal{W}:=\{w_i|i=1,\dots,N$\}; Number of samples $S$}%
	\KwOutput{Set of samples $\{(\vec{\theta}_i,y_i)|i=1,\ldots,S\}$}
	\For{$i=1$ \KwTo $S$}{
		$y_i$ $\leftarrow$ success\;
		$\vec{\theta}_i \leftarrow$ SampleRandomDisplacement()\;
		$\mat{T}^\textit{sensor}_\textit{marker}$ $\leftarrow$ DetectMarker($\mathcal{W}$)\;
		$\mat{T}^\textit{world}_\textit{sensor}$ $\leftarrow$ ComputeForwardKinematics()\;
		\tcp{\small{end effector pose in object (marker) coordinate system}}
		$\hat{\mat{T}}^\textit{marker}_\textit{grasp}$ $\leftarrow$ SamplePose($\vec{\theta}_i, \mat{T}^\textit{marker}_\textit{grasp}$)\;
		\tcp{\small{end effector pose in world coordinate system}}
		$\hat{\mat{T}}^\textit{world}_\textit{grasp}$ $\leftarrow$  $\mat{T}^\textit{world}_\textit{sensor} \cdot \mat{T}^\textit{sensor}_\textit{marker} \cdot \hat{\mat{T}}^\textit{marker}_\textit{grasp}$\;
		$\text{GraspObject}(\hat{\mat{T}}^\textit{world}_\textit{grasp}, \mathcal{W})$\;
		\If{\text{SuccessfulGrasp}() is False} 
		{
			\tcp{\small{marker detected on the table or force limits exceeded}}
			$y_i$ $\leftarrow$ failure \\
		}\Else{
			$\text{InstallObject}(\mathcal{W})$\;

			\If{$\text{SuccessfulInstall}()$ is False} 
			{
				\tcp{\small{marker on wrong pose or force limits exceeded}}
					$y_i$ $\leftarrow$ failure
		}}
		Record($\vec{\theta}_i$, $y_i$)\;
		MoveObjectToStart($\mathcal{W})$;
	}
	\caption{Practical sampling of the pose space}\label{alg:sampling}
\end{algorithm}
\section{Experiments}
\label{sec:experiments}
We implemented four assembly tasks for a robot arm. For each task the gripper and custom made
fingers were used.
To accurately estimate the success probabilities,
a large number of pose samples were needed for each task. For that reason, the setups were made autonomous
so that the task success was automatically detected.
This was achieved by verifying the final pose of the assembled parts and measuring the torque sensor readings during
the task execution (Section~\ref{sec:sampling}).
The tasks,
robot setups, experimental results and verification experiments are explained
in the following.

\subsection{Tasks}
To conduct experiments on practical tasks they were
selected from the production line of a local engine manufacturing company.
The selected tasks were: (Task 1) installation of a motor cap 1, (Task 2) installation of a motor frame and (Task 3) installation of a motor cap 2
(different engine model).
The fourth task
(Task 4) is different from others: picking and dropping a part to a container (the {\em faceplate} part from the Cranfield assembly benchmark). 
As Task 4 does not require precise manipulation, the task requires less accurate pose than the
others.
This can be verified in Table~\ref{tab:limits} where
the Task 4 limits are less strict (by order of magnitude) as compared to the other tasks.
Cranfield faceplate was selected since its 3D model is
publicly available and the part is used in robot manipulation studies.
The tasks were programmed by an experienced engineer who also carefully selected the grippers and fingers. The engineer was instructed that accurate pose is always available.

\subsection{Setup}
\begin{figure}[h]
	\begin{center}
		\includegraphics[width=.99\linewidth]{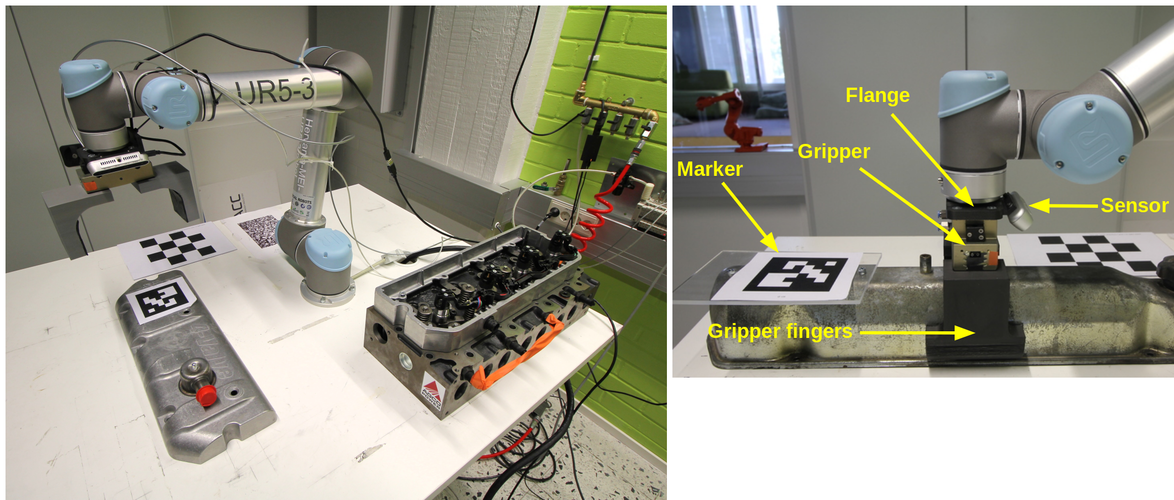}  
		\caption{The experimental robot setup to sample the pose space of the engine cap 1. The task is to grasp and accurate assembly the cap to the engine block. Failures in task execution were automatically detected during sampling (Section~\ref{sec:sampling}).
			\label{fig:setup}}
	\end{center}
\end{figure}

In Fig.~\ref{fig:setup} is illustrated the robotic setup used in our experiments.
The setup consisted of a model 5
Universal Robot Arm (UR5) and a Schunk PGN-100 gripper.
The gripper operates pneumatically and was configured to have a high gripping force (approximately 600N) to prevent object slippage.
In addition, the gripper had custom 3D printed fingers plated with rubber.
For visual perception, an Intel RealSense D415 RGB-D sensor was secured on a 3D printed flange mounted between the gripper and the robot end effector. 
All the in-house made 3D prints were made using nylon reinforced with carbon fiber to tolerate external forces during the experiments.
The computation was performed on a single laptop with Ubuntu 18.04.
All tasks and the canonical grasp poses were validated by executing the task 100 times with pose obtained using the 2D patterns (Section~\ref{sec:sampling}). No failures occurred during the validation. On average, successful executions took 45-55 seconds and in 24 hours the robot was able to execute approximately 1,100 attempts.
The setup was able automatically to recover from most of the failure cases (dropping the object, object collision, etc.), however, if the marker was occluded by the environment or if the manipulated object got jammed against internal parts of the motor, the system was restarted by a human operator.

\subsection{RGB-D dataset}
\begin{figure}[h]
	\begin{center}
		\includegraphics[width=1.0\linewidth]{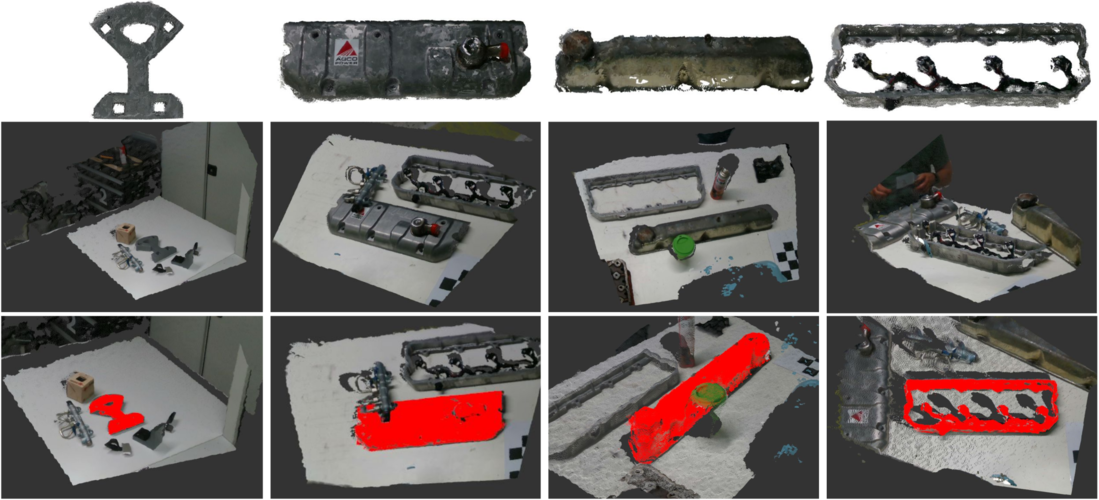}  
		\caption{Top: Point cloud models of the used industry objects; faceplate, motor cap 1, motor cap 2 and a motor frame. The models were reconstructed by combining different view points of the robot arm and RGB-D sensor. Middle: example test samples (colored point clouds). Bottow: renderings of the object models on the test images using the ground truth poses.  
		\label{fig:object_models}}
	\end{center}
\end{figure}

In the dataset each of the object models are stored as a point cloud that represents a set of N 3D points $\left\{\vec{x}_i|i=1,\ldots,N\right\}$.
In addition, for each point the corresponding color value $\vec{c} \in \mathbb{N}^3$ is stored.
The models were generated using the same setup. The point cloud models were
obtained automatically by moving the robot arm with the attached
RGB-D sensor around each object. By using the camera poses obtained
from robot kinematics the measurements were merged to a single point
cloud that is the stored object model (see Fig. \ref{fig:object_models}). %
The automatically captured point clouds were then manually checked and
all artifacts and redundant parts of the reconstructed point cloud were removed manually using the open-source mesh processing software MeshLab~\cite{cignoni2008meshlab}.
Finally, the coordinate system of each model point cloud was aligned with the canonical grasp pose of the object. 

The test dataset was generated in a similar manner by moving the arm around the objects. For each of the objects 150 test images were collected in three different settings: 1) a single target object present, 2) multiple objects present and 3) the target object partially occluded by other object(s). The dataset contains manually verified ground truth to align the model point cloud to each test image and further to locate the canonical grasp pose relative to sensor ($\mat{T}_{\textit{sensor}}^{\textit{grasp}}$).

\subsection{Model validation}
The probability model $P(X=1|\vec{\theta})$ in Section~\ref{sec:probability}
was fitted using the sampling procedure in Section~\ref{sec:sampling}.
For all tasks approximately 3,300 valid samples were generated around task canonical poses.

\begin{figure}[t]
	\centering
	\includegraphics[width=0.99\linewidth]{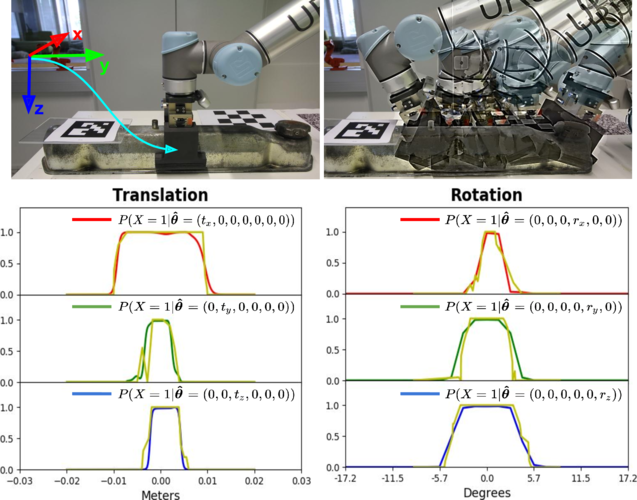}
	\caption{Motor cap 2 used in our Task 3. The coordinate system is object centric (top left) and pose samples are taken around a canonical grasp pose (see experiments for more details). Below are the estimated (the red, green and blue lines) and validated success probabilities (yellow line) on the six main axes (three translations and three rotations) in vicinity of the canonical grasp pose.
		\label{fig:pull_figure}}
\end{figure}

\begin{table}[]
	\caption{Sampling limits for translation $(t_x,t_y,t_z)$ and rotation $(r_x,r_y,r_z)$ in meters and degrees, respectively. Beyond these limits the task always fails. %
	}
	\label{tab:limits}
	\centering
\resizebox{0.99\linewidth}{!}{%
	\begin{tabular}{lllllllll}
		\toprule
		{\em Variable} &
		\multicolumn{8}{c}{\em Task Name}\\
		& \multicolumn{2}{c}{Task 1} & \multicolumn{2}{c}{Task 2} & \multicolumn{2}{c}{Task 3} & \multicolumn{2}{c}{Task 4} \\ \midrule
		$t_x$ & \multicolumn{2}{l}{{[}-9.0, 9.0{]} $\cdot 10^{-3}$}                 & \multicolumn{2}{l}{{[}-6.0, 6.0{]} $\cdot 10^{-3}$} & \multicolumn{2}{l}{{[}-9.0, 9.0{]} $\cdot 10^{-3}$}  & \multicolumn{2}{l}{{[}-6.5, 8.5{]} $\cdot 10^{-3}$} \\ 
		$t_y$ & \multicolumn{2}{l}{{[}-1.0, 1.0{]} $\cdot 10^{-3}$}                 & \multicolumn{2}{l}{{[}-3.0, 2.5{]} $\cdot 10^{-3}$} & \multicolumn{2}{l}{{[}-5.0, 6.0{]} $\cdot 10^{-3}$}  & \multicolumn{2}{l}{{[}-2.1, 2.1{]} $\cdot 10^{-2}$} \\ 
		$t_z$ & \multicolumn{2}{l}{{[}-1.0, 5.0{]} $\cdot 10^{-3}$}                 & \multicolumn{2}{l}{{[}-2.0, 4.0{]} $\cdot 10^{-3}$} & \multicolumn{2}{l}{{[}-2.0, 5.0{]} $\cdot 10^{-3}$}  & \multicolumn{2}{l}{{[}-1.2, 1.7{]} $\cdot 10^{-2}$} \\ 
		$r_x$  & \multicolumn{2}{l}{{[}-6.3, 6.3{]} $\cdot 10^{0}$}       & \multicolumn{2}{l}{{[}-6.3, 6.3{]} $\cdot 10^{0}$} & \multicolumn{2}{l}{{[}-2.0, 1.0{]} $\cdot 10^{0}$}    & \multicolumn{2}{l}{{[}-1.5, 1.5{]} $\cdot 10^{1}$} \\ 
		$r_y$  & \multicolumn{2}{l}{{[}-5.0, 5.0{]} $\cdot 10^{-1}$}       & \multicolumn{2}{l}{{[}-2.5, 1.0{]} $\cdot 10^{0}$} &  \multicolumn{2}{l}{{[}-2.0, 2.0{]} $\cdot 10^{0}$}   & \multicolumn{2}{l}{{[}-1.5, 1.5{]} $\cdot 10^{1}$} \\ 
		$r_z$  & \multicolumn{2}{l}{{[}-5.0, 5.0{]} $\cdot 10^{-1}$}       & \multicolumn{2}{l}{{[}-1.5, 1.5{]} $\cdot 10^{0}$} & \multicolumn{2}{l}{{[}-4.0, 4.0{]} $\cdot 10^{0}$}    & \multicolumn{2}{l}{{[}-1.5, 1.5{]} $\cdot 10^{1}$} \\ 
		\bottomrule
	\end{tabular}}
\end{table}
The estimated probability models were validated by sampling each dimension separately on grid points and executing the task ten times on each point with real robot. 
The averaged task success rate on real robot was then compared against the proposed models and the estimated probabilities matched well as can be seen in Fig.~\ref{fig:pull_figure}.
\subsection{Methods}
Comparison included the methods in
Section~\ref{sec:SotA}. All methods input point clouds
of the model and scene. The model and scene point clouds were downsampled to fixed resolutions using a regular voxel grid to limit the amount of data for processing. Depending on the density of the cloud the size of voxels was $0.5-1.0mm$. Since methods also exploit surface normals they were
estimated using the standard least squares plane fitting on points in a small neighborhood. 
To further reduce the computational complexity in the matching stage, 
we avoided using all the surface points as local keypoints and only select a uniform subset of $1000-3000$ points per object model using the voxel grid filtering. 
Finally, local descriptors for point matching were computed using the
local point neighborhoods. The SHOT~\cite{tombari2010unique} feature descriptor was selected since it performed the best in the preliminary experiments. The descriptor support radius was set to
$0.125\times$ the object model's minimal bounding box diagonal.
For each test scene, the best matching descriptors in $L_2$ sense between the model and scene were selected using a randomized $kd$-tree similarity search. The best matches formed then the initial set of correspondences
for each method.

\subsection{Performance indicators}

\begin{figure}[t]
	\centering
	\includegraphics[width=0.6\linewidth]{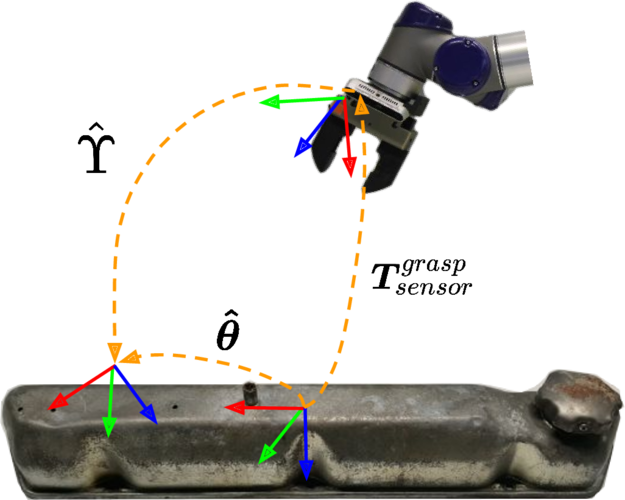}
	\caption{Coordinate frames in the evaluation procedure.
		\label{fig:eval_description}}
\end{figure}

The main performance metric in our work is the estimated success
probability defined in Section~\ref{sec:probability}. The probabilities
were computed around the canonical grasp pose of each object
and therefore the sampled values actually represent residual from this pose (see Fig. \ref{fig:eval_description}). The corresponding object-relative grasp pose of the pose estimate $\mat{\hat{\Upsilon}}$ is calculated as:
\begin{equation}
\vec{\hat{\theta}} = \mat{\Phi}^{-1}\big(\mat{T}_{\textit{sensor}}^{\textit{grasp}} \mat{\hat{\Upsilon}}\big) \enspace,
\end{equation}
where the transformation matrices $\mat{T}_{\textit{sensor}}^{\textit{grasp}}$
defines the canonical grasp pose respect to the sensor coordinate system.
The $\mat{\Phi}^{-1}(\cdot)$ operator converts the $4\times4$ pose matrix to 6D vector representation.  
Finally, the task success is evaluated using the proposed metric as $P(X\!=\!1 | \vec{\hat{\theta}})$.
We calculated the average probabilities over the whole
dataset and also the proportion of images for which the probability
is greater or equal to $0.90$.

In addition to the proposed indicator we also report the ADC error calculated over the points transformed by 
the ground truth and estimated object pose as suggested in~\cite{hinterstoisser2012model}. The ADC error is computed from
\begin{equation}
\vec{\epsilon}_{ADC} = \frac{1}{|\cal{M}|}\sum_{\vec{x}\in\cal{M}} \left\Vert \mat{\hat{\Upsilon}}\vec{x} - \mat{\Upsilon}\vec{x}\right\Vert 
\end{equation}
where $\cal{M}$ is the set of model 3D points. We also
report the top$-25\%$ ADC error, which is less affected by
outliers.

\subsection{Results}

\begin{table*}[t]
	\caption{Comparison of pose estimation methods with our dataset (single: single object in the scene; multi: multiple objects (clutter); occ: multiple objects and occlusion: all: average over all test samples).%
	}
	\label{tab:comparison1}
	\centering
	\resizebox{1.0\linewidth}{!}{%
		\begin{tabular}{l  rrrrrrr rrrrrrr}
			\toprule
		    & \multicolumn{7}{c}{Task: {\em Task 1}} & \multicolumn{7}{c}{Task: {\em Task 2}}  \\
			& \multicolumn{7}{c}{Part: {\em Motor cap 1}; Gripper: {\em Shunker}} & \multicolumn{7}{c}{Part: {\em Motor frame}; Gripper: {\em Shunker}} \\
			& \multicolumn{7}{c}{Fingers: {\em Custom made}} & \multicolumn{7}{c}{Fingers: {\em Custom made}} \\
			&  \multicolumn{7}{c}{\includegraphics[width=3.5cm]{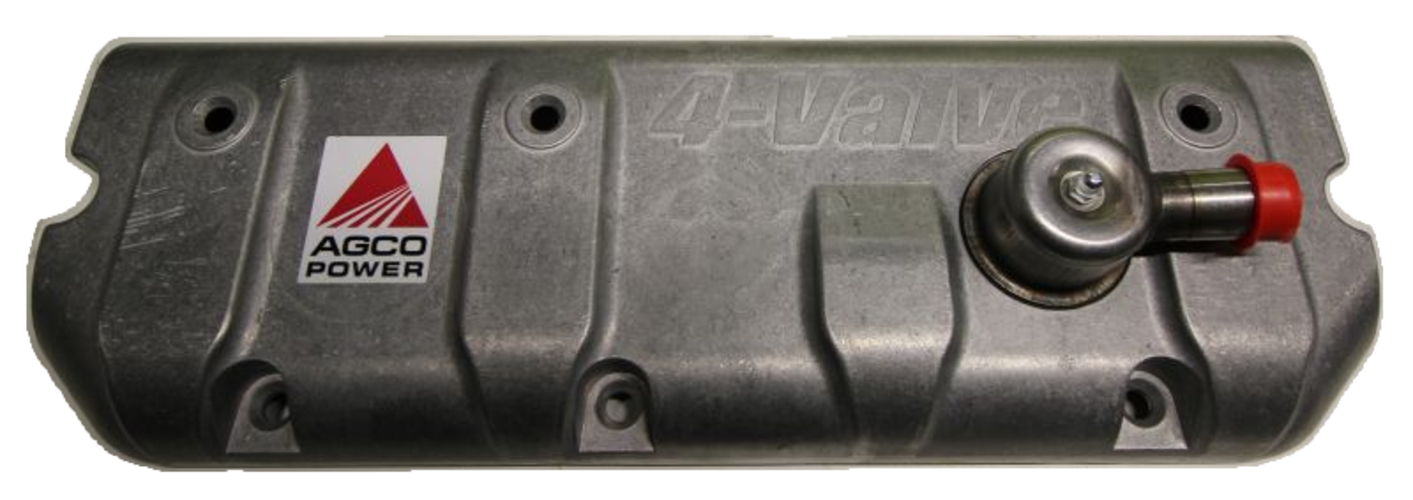}\hspace{\medskipamount}\includegraphics[width=2.0cm]{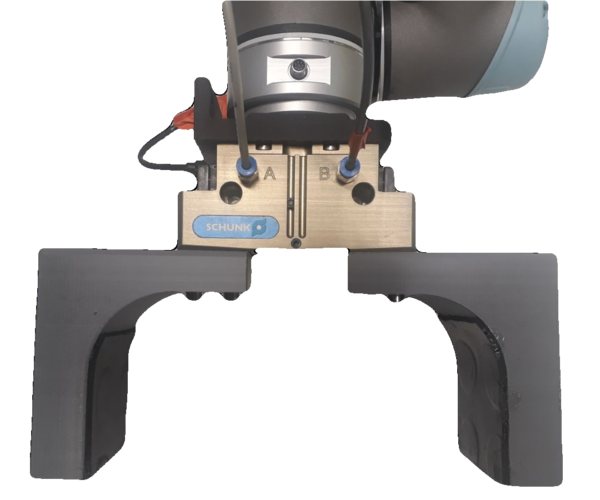}}
			& \multicolumn{7}{c}{\includegraphics[width=3.5cm]{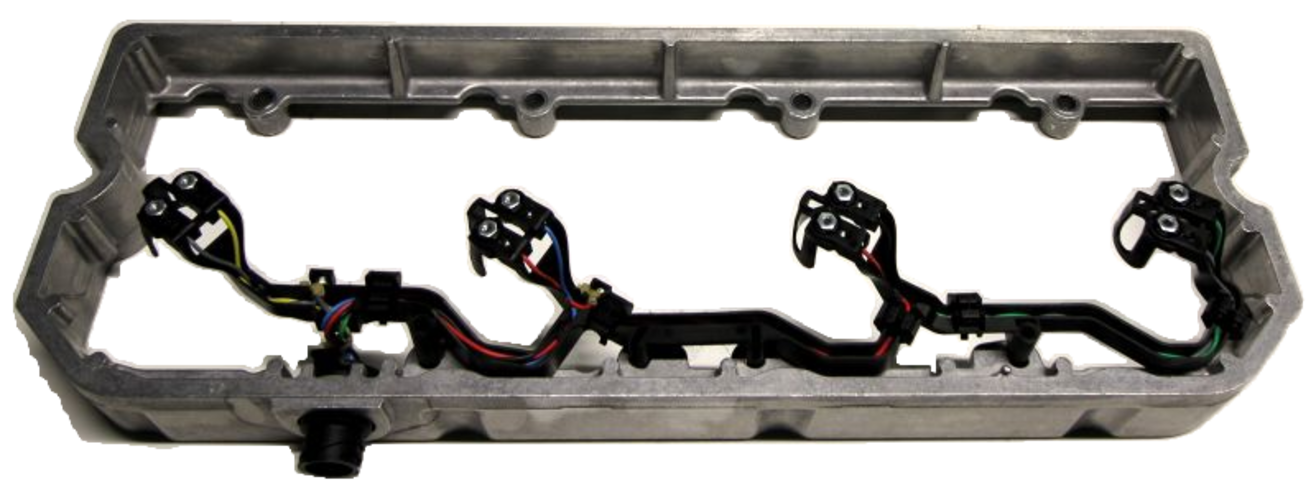}\hspace{\medskipamount}\includegraphics[width=2.0cm]{./resources/gripper.png}} \\
			\cmidrule(r){2-8}\cmidrule(r){9-15} 
			\multirow{ 2}{*}{Method} & \multicolumn{4}{c}{Average success probability} & \multicolumn{1}{c}{$\%\{ p\ge0.9 \}$} & \multicolumn{2}{c}{Avg. ADC} & \multicolumn{4}{c}{Average success probability} & \multicolumn{1}{c}{$\%\{ p\ge0.9 \}$} & \multicolumn{2}{c}{Avg. ADC} \\
			\cmidrule(r){2-5}\cmidrule(r){6-6}\cmidrule(r){7-8}\cmidrule(r){9-12}\cmidrule(r){13-13}\cmidrule(r){14-15} 
			& single & multi & occ & all & all & all & best-25\% & single & multi & occ & all & all &  all & best-25\% \\
			\cmidrule(r){1-8}\cmidrule(r){9-15} 
			GC~\cite{GC}  						   
			& 0.24 & 0.18 & 0.12 & 0.19 & 12\% & 0.08 & 3.83$\cdot10^{-3}$  & 0.21 & 0.22 & 0.19 & 0.21 &  9\% & 0.02 & 5.36$\cdot10^{-3}$\\
			HG~\cite{tombari2010object}            
			& 0.31 & 0.29 & 0.20 & 0.26 & 14\% & 0.06 & 3.87$\cdot10^{-3}$ & 0.28 & 0.27 & 0.27 & 0.28 & 15\% & 0.03 & 5.19$\cdot10^{-3}$\\
			SI~\cite{glent2014search}              
			& 0.00 & 0.00 & 0.00 & 0.00 & 0\% & 0.46 & 1.78$\cdot10^{-1}$ & 0.14 & 0.04 & 0.03 & 0.07 &  5\% & 0.42 & 1.81$\cdot10^{-2}$\\
			ST~\cite{leordeanu2005spectral}              
			& 0.01 & 0.03 & 0.00 & 0.01 & 0\% & 0.35 & 9.12$\cdot10^{-2}$ & 0.23 & 0.16 & 0.07 & 0.17 & 7\% & 0.34 & 4.38$\cdot10^{-3}$\\
			NNSR~\cite{hough1962method}            
			& 0.00 & 0.00 & 0.00 & 0.00 & 0\% & 0.26 & 1.18$\cdot10^{-1}$ & 0.00 & 0.00 & 0.00 & 0.00 &     0\% & 0.36 & 1.50$\cdot10^{-1}$\\
			RANSAC~\cite{fischler1981random}       
			& 0.00 & 0.00 & 0.00 & 0.00 & 0\% & 0.75 & 1.71$\cdot10^{-1}$ & 0.00 & 0.00 & 0.00 & 0.00 &     0\% & 0.65 & 2.03$\cdot10^{-1}$\\
			\bottomrule
			\toprule
			& \multicolumn{7}{c}{Task: {\em Task 3}} & \multicolumn{7}{c}{Task: {\em Task 4}}  \\
			& \multicolumn{7}{c}{Part: {\em Motor cap 2}; Gripper: {\em Shunker}} & \multicolumn{7}{c}{Part: {\em Cranfield faceplate}; Gripper: {\em Shunker}} \\
			& \multicolumn{7}{c}{Fingers: {\em Custom made}} & \multicolumn{7}{c}{Fingers: {\em Custom made}} \\
			& \multicolumn{7}{c}{\includegraphics[width=3.5cm]{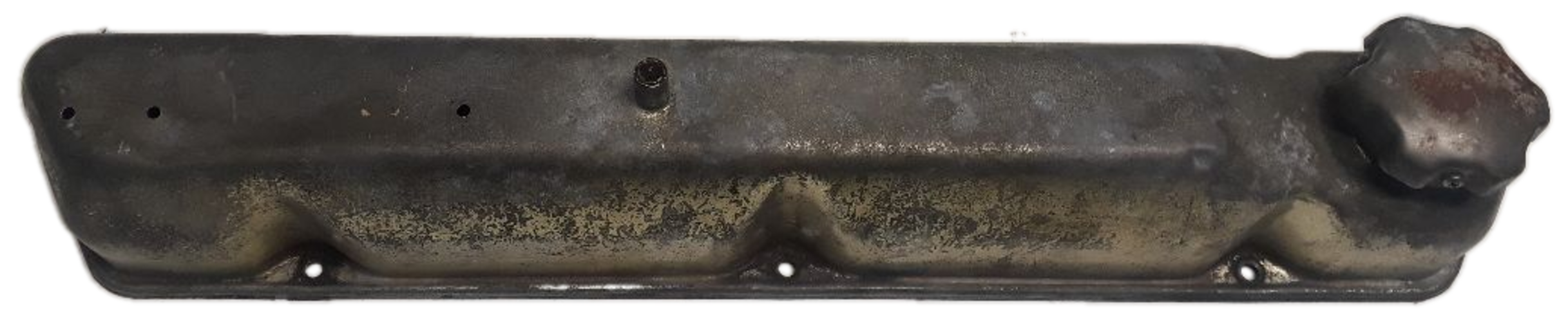}\hspace{\medskipamount}\includegraphics[width=2.0cm]{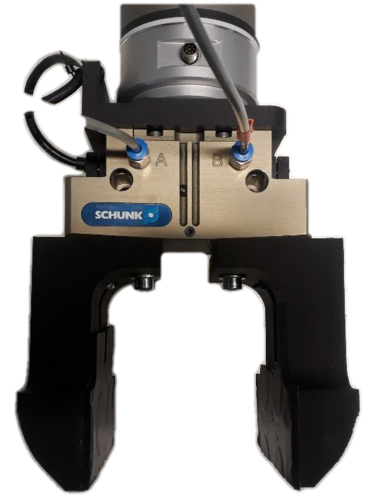}}
			& \multicolumn{7}{c}{\includegraphics[width=2cm]{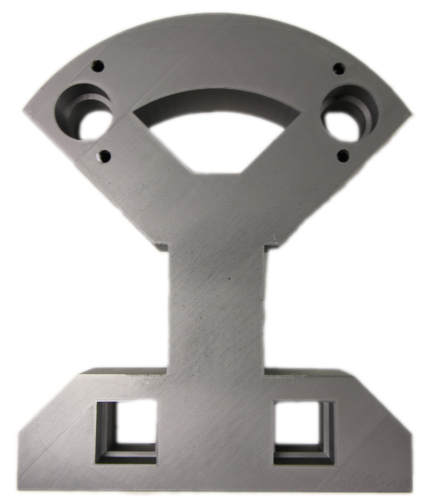}\hspace{\medskipamount}\includegraphics[width=2cm]{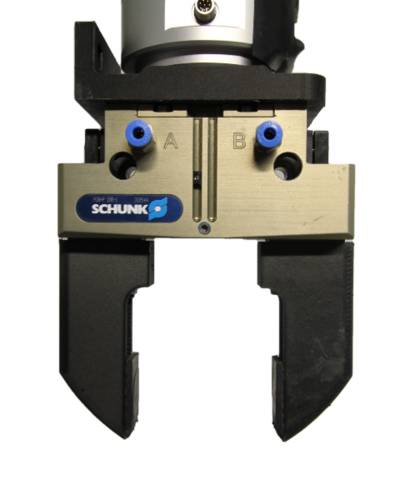}} \\
			\cmidrule(r){2-8}\cmidrule(r){9-15} 
			\multirow{ 2}{*}{Method} & \multicolumn{4}{c}{Average success probability} & \multicolumn{1}{c}{$\%\{ p\ge0.9 \}$} & \multicolumn{2}{c}{Avg. ADC} & \multicolumn{4}{c}{Average success probability} & \multicolumn{1}{c}{$\%\{ p\ge0.9 \}$} & \multicolumn{2}{c}{Avg. ADC} \\
			\cmidrule(r){2-5}\cmidrule(r){6-6}\cmidrule(r){7-8}\cmidrule(r){9-12}\cmidrule(r){13-13}\cmidrule(r){14-15} 
			& single & multi & occ & all & all & all & best-25\% & single & multi & occ & all & all &  all & best-25\% \\
			\cmidrule(r){1-8}\cmidrule(r){9-15} 
			GC~\cite{GC}  						   
			& 0.24 & 0.25 & 0.20 & 0.24 & 13\% & 0.09 & 6.28$\cdot10^{-3}$ & 0.66 & 0.67 & 0.59 & 0.64 & 65\% & 0.15 & 4.57$\cdot10^{-3}$ \\
			HG~\cite{tombari2010object}            
			& 0.13 & 0.21 & 0.10 & 0.15 &  9\% & 0.11 &  7.81$\cdot10^{-3}$ & 0.64 & 0.68 & 0.56 & 0.63 & 60\% & 0.16 & 3.43$\cdot10^{-3}$ \\
			SI~\cite{glent2014search}              
			& 0.11 & 0.19 & 0.11 & 0.13 &  8\% & 0.09 & 1.11$\cdot10^{-2}$ & 0.37 & 0.43 & 0.20 & 0.35 & 35\% & 0.39 & 9.94$\cdot10^{-3}$\\
			ST~\cite{leordeanu2005spectral}              
			& 0.17 & 0.18 & 0.08 & 0.15 & 8\% & 0.11 & 5.46$\cdot10^{-3}$ & 0.40 & 0.39 & 0.30 & 0.37 & 36\% & 0.30  & 6.47$\cdot10^{-3}$ \\
			NNSR~\cite{hough1962method}            
			& 0.02 & 0.00 & 0.00 & 0.01 &  1\% & 0.19 & 6.10$\cdot10^{-2}$ & 0.05 & 0.04 & 0.07 & 0.05 &  5\% & 0.28 & 7.16$\cdot10^{-2}$ \\
			RANSAC~\cite{fischler1981random}       
			& 0.00 & 0.00 & 0.00 & 0.00 &  0\% & 0.28 & 1.24$\cdot10^{-1}$ & 0.00 & 0.04 & 0.00 & 0.01 &  1\% & 0.51 & 1.05$\cdot10^{-1}$ \\
			\bottomrule
		\end{tabular}
	}
\end{table*}

The results for all methods and parts are in Table \ref{tab:comparison1}.
The two best methods are Hough Transform (HG) by
Tombari et al.~\cite{tombari2010object} and GC by
Chen and Bhanu~\cite{GC}. HG and GC perform considerably better than
the two more state-of-the-art methods SI and ST although the performance
of all methods remains surprisingly low. The two baselines, simple
Hough voting (NNSR) and RANSAC, perform poorly.

\begin{figure}[h]
	\begin{center}
		\includegraphics[width=1.0\linewidth]{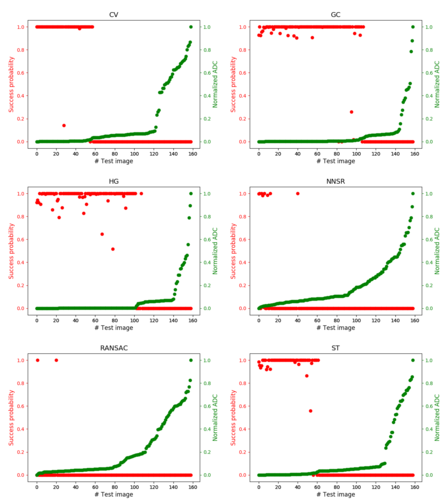}  
		\caption{ADC pose error (green) and success probability (red) of all test images of Task 4 for different pose estimation methods. 
Images are sorted based on their ADC error. 
Note rapid change from success (1.0) to failure (0.0) when the error goes beyond certain points.
\label{fig:adc_vs_gp}}
	\end{center}
\end{figure}

\begin{figure}[h]
	\begin{center}
		\includegraphics[width=1.0\linewidth]{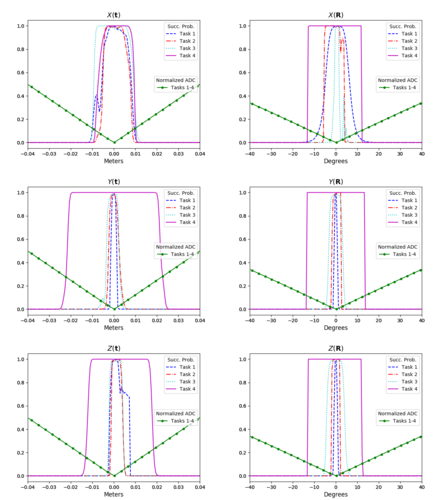}
		\caption{ADC and success probability from
		controlled experiments for all the tasks (Task 1 -- Task 4). Effect of
		rotation (left column) and translation (right column) to
		the ADC and success probability.}
		\label{fig:controlled_adc_vs_gp}
	\end{center}
\end{figure}

\begin{figure}[h]
	\begin{center}
		\includegraphics[width=1.0\linewidth]{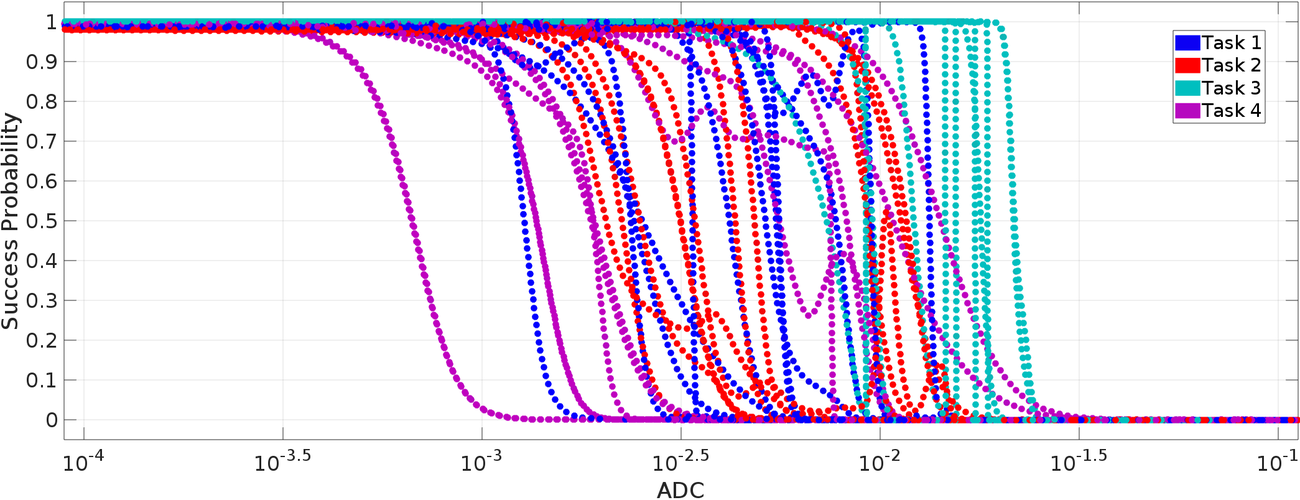}
		\caption{Success Probability vs. ADC scatterplot from
		controlled experiments for all the tasks (Task 1 -- Task 4). The scatterplot shows that the ADC does not reflect the success probability, except for extreme and trivial cases of failure or success; the two measures cannot be put in correspondence to each other not even through a nonlinear mapping.
		\label{fig:succ_adc_xyplot}}
	\end{center}
\end{figure}

\paragraph{Success probability vs.~ADC}
It is important to notice
that the ADC results indicate clearly smaller difference between the
methods than indicated by the success probability. The
success probability measures directly performance in
the physical task. This is even more evident in
Fig.~\ref{fig:adc_vs_gp} of the ADC error
and success probability graphs. The success probability
is able to measure the points after which the success
quickly drops from $1.0$ to $0.0$, but ADC
(green points) regrades linearly even after these
points and is thus uninformative.
The non-linear behavior was verified in the controlled experiments in all the tasks as illustrated in
Fig.~\ref{fig:controlled_adc_vs_gp}.
Moreover, the difference between the two metrics is further illustrated in the success probability vs. ADC scatter plots of all four tasks in Fig.~\ref{fig:succ_adc_xyplot}.

\section{Conclusions}
This work addressed evaluation of vision based object pose estimation methods for robotics. In our experiments we demonstrated how the popular error measure, ADC, poorly indicates success in robot manipulation
tasks and is therefore uninformative.
As a novel solution, we proposed a probabilistic metric that measures the true success rate without the physical setup. The experimental results demonstrated poor
performance of the existing methods which indicates that more work is needed for 6D object pose estimation in robotics. 
All data and code will be made publicly available
to facilitate fair comparisons and to boost research
on robot vision for vision based object grasping and manipulation.

\section*{References}

\bibliography{robotics_pose}

\end{document}